\PassOptionsToPackage{table}{xcolor}
\documentclass[lettersize,journal]{IEEEtran}

\usepackage{tikz,pgfplots}
\usepackage{color}
\usepackage[table]{xcolor}
\newcommand{\revisao}[1]{{#1\normalfont}}
\usepackage{amsmath}
\usepackage{hyperref} 
\usepackage{aas_macros}
\def\checkmark{\tikz\fill[scale=0.4](0,.35) -- (.25,0) -- (1,.7) -- (.25,.15) -- cycle;}

\begin{document}
\title{Assessing the Capability of LLMs in Solving POSCOMP Questions}

\author{
    \IEEEauthorblockN{
        Cayo Viegas\IEEEauthorrefmark{3}, Rohit Gheyi\IEEEauthorrefmark{3}, Márcio Ribeiro\IEEEauthorrefmark{2}\\
    }
    \IEEEauthorblockA{
        \IEEEauthorrefmark{3}UFCG, Brazil\\ cayo.viegas@ccc.ufcg.edu.br and rohit@dsc.ufcg.edu.br\\
    }
    \IEEEauthorblockA{
        \IEEEauthorrefmark{2}UFAL, Brazil\\ marcio@ic.ufal.br\\
    }
}

\markboth{Journal of \LaTeX\ Class Files,~Vol.~14, No.~8, August~2021}%
{Shell \MakeLowercase{\textit{et al.}}: A Sample Article Using IEEEtran.cls for IEEE Journals}

\maketitle

\begin{abstract}
Recent advancements in Large Language Models (LLMs) have significantly expanded the capabilities of artificial intelligence in natural language processing tasks.
Despite this progress, their performance in specialized domains such as computer science remains relatively unexplored.
Understanding the proficiency of LLMs in these domains is critical for evaluating their practical utility and guiding future developments. The POSCOMP, a prestigious Brazilian examination used for graduate admissions in computer science promoted by the Brazlian Computer Society (SBC), provides a challenging benchmark.
This study investigates whether LLMs can match or surpass human performance on the POSCOMP exam. Four LLMs -- ChatGPT-4, Gemini 1.0 Advanced, Claude 3 Sonnet, and Le Chat Mistral Large -- were \revisao{initially} evaluated on the 2022 and 2023 POSCOMP exams.
The assessments measured the models' proficiency in handling complex questions typical of the exam. LLM performance was notably better on text-based questions than on image interpretation tasks. In the 2022 exam, ChatGPT-4 led with 57 correct answers out of \revisao{69} questions, followed by Gemini 1.0 Advanced (49), Le Chat Mistral (48), and Claude 3 Sonnet (44). Similar trends were observed in the 2023 exam.
ChatGPT-4 achieved the highest performance, surpassing all students who took the POSCOMP 2023 exam.
LLMs, particularly ChatGPT-4, show promise in text-based tasks on the POSCOMP exam, although image interpretation remains a challenge. 
\revisao{Given the rapid evolution of LLMs, we expanded our analysis to include more recent models -- o1, Gemini 2.5 Pro, Claude 3.7 Sonnet, and o3-mini-high -- evaluated on the 2022–2024 POSCOMP exams. These newer models demonstrate further improvements and consistently surpass both the average and top-performing human participants across all three years.}
\end{abstract}


\section{Introduction}
\label{sec:introducao}

\revisao{\textit{Context.}} 
The POSCOMP~\cite{poscomp} is a prestigious assessment designed to test the knowledge of prospective computer science graduate students, promoted by the Brazilian Computer Society (SBC). It serves as an entry criterion for many graduate programs across Brazil. Using this exam as a benchmark for evaluating Large Language Models (LLMs) allows for a direct comparison between AI capabilities and human standards, offering valuable insights into the strengths and limitations of current AI models.

Recent advancements in LLMs~\cite{Goodfellow-et-al-2016,attention-is-all-you-need} have significantly expanded the capabilities of Artificial Intelligence (AI), particularly in natural language processing tasks. These developments have prompted increasing interest in understanding how well LLMs can solve complex problems in specialized domains, such as computer science.

Recent studies have examined the performance of LLMs on Brazilian educational assessments. 
\revisao{
Mendon\c{c}a~\cite{DBLP:journals/toce/Mendonca24} evaluated ChatGPT-4 Vision on Brazil's undergraduate Computer Science ENADE exam, showing that the model outperformed average human participants, despite struggling with certain aspects of visual reasoning. 
}
Similarly, Pires et al.~\cite{pires} assessed ChatGPT-4 Vision and ChatGPT-4 Text on the national ENEM exams, finding that text-only prompts with image captions outperformed image-only ones when using few-shot prompting with Chain-of-Thought~\cite{chain-of-thought}. Nunes et al.~\cite{nunes} conducted a comprehensive analysis of ChatGPT-3.5 and ChatGPT-4 on ENEM questions from 2009–2017 and 2022, revealing that ChatGPT-4 with Chain-of-Thought achieved an impressive 87\% accuracy on the 2022 exam.

While these studies demonstrate promising results for LLMs in high school and undergraduate-level assessments, little is known about their performance on graduate-level assessments such as POSCOMP. This gap motivates a deeper investigation into how well state-of-the-art LLMs perform in this more advanced context.

\revisao{\textit{Study Objectives.}}
In this article, we investigate the ability of state-of-the-art LLMs to solve questions from the POSCOMP exam. We evaluate the performance of four models -- ChatGPT-4~\cite{gpt}, Gemini 1.0 Advanced~\cite{gemini}, Claude 3 Sonnet~\cite{claude}, and Le Chat Mistral Large~\cite{mistral} -- on the 2022 and 2023 editions of the exam, aiming to determine whether these models can match or surpass human performance in a rigorous academic setting.
\revisao{In light of the rapid evolution of LLMs, we extend this analysis to include more recent frontier models, such as o1, Gemini 2.5 Pro Experimental, Claude 3.7 Sonnet, and o3-mini-high. This expanded evaluation includes the 2024 edition of the POSCOMP and allows us to assess the progress of cutting-edge LLMs across three consecutive years.}

\revisao{\textit{Main Contributions.}}
Our findings show that ChatGPT-4 consistently outperformed the other models, achieving the highest scores in 2023 and demonstrating particularly strong performance in Mathematics and Computer Science Fundamentals. While Gemini 1.0 and Le Chat Mistral performed reasonably well across various subjects, only ChatGPT-4 exhibited consistent excellence. Claude 3 Sonnet showed strength in explanation and reasoning tasks but struggled in quantitative areas.

\revisao{
We also observed significant performance gains in more recent models. Gemini 2.5 Pro Experimental showed substantial improvement over Gemini 1.0, and, along with o3-mini-high, achieved or exceeded 90\% accuracy across several topics -- surpassing both average and top human participants. In addition to performance benchmarking, our study highlights the practical utility of LLMs in educational assessment design. By examining model responses, problem setters can identify ambiguities, clarify question wording, and refine answer choices -- ultimately improving the quality and fairness of exam content. 
}

The remainder of the paper is organized as follows. 
\revisao{Section~\ref{sec:related} relates our work to previous ones.}
Section~\ref{sec:poscomp} provides details about the POSCOMP exam. Section~\ref{sec:evaluation} presents evaluations of LLM performance in answering POSCOMP questions. Section~\ref{sec:threats} outlines the threats to the validity of the study.
Finally, Section~\ref{sec:conclusao} concludes our work.

\revisao{
\section{Related Work}
\label{sec:related}
}

Pires et al.~\cite{pires} evaluated ChatGPT-4 Vision and ChatGPT-4 Text in two Brazilian ENEM exams. Text-only prompts with captions describing images outperformed the image-only version. They used few-shot prompting with Chain-of-Thought~\cite{chain-of-thought}. The LLM struggled most in the Mathematics section.
In our work, we evaluated four LLMs on two POSCOMP editions using zero-shot prompts. Additionally, ChatGPT-4 performed well on math questions.

Zhang et al.~\cite{zhang} assessed ChatGPT-3.5, ChatGPT-4, and ERNIE-Bot (including its Turbo version) with questions from the 2010-2022 Chinese University Entrance Exam (GAOKAO), excluding those containing images. Zero-shot prompting and human evaluation were used. Results showed that the models performed well on knowledge-based questions but struggled with specific logical reasoning, math problems, and reading comprehension of longer Chinese texts. 
In our work, we evaluated four LLMs on two POSCOMP editions using zero-shot prompts, where ChatGPT-4 excelled in math.

Guillen-Grima et al.~\cite{grima} examined ChatGPT-3.5 and ChatGPT-4's performance on the 2022 Spanish Medical Residency Entrance Exam, in both the native language and English translations, finding a slightly better performance in English. 
In our study, we assessed four LLMs on two POSCOMP editions. After using image prompts in Portuguese, we translated the questions into English and saw improved performance across all models.

Bommarito and Katz~\cite{bommaritoM} evaluated ChatGPT-3.5 on the Multistate Bar Examination portion of the National Conference of Bar Examiners' Bar Exam, a prerequisite for legal practice in the U.S. ChatGPT-3.5 achieved a passing rate and scores comparable to human examinees. 
In our evaluation, two of the models tested in POSCOMP 2023 scored higher than human examinees.

Bommarito et al.~\cite{BommaritoJ} tested ChatGPT-3.5 on two assessments based on the American Institute of Certified Public Accountants' Uniform CPA Exam Blueprints, a U.S. accounting exam. In the first test, which involved quantitative reasoning, the model using text-davinci-003 achieved a modest 14.4\%{} accuracy. The second assessment, focused on fundamental skills without quantitative reasoning, reached 57\%{} accuracy, significantly above chance and close to reported human performance. 
In our study, we assessed four LLMs on two POSCOMP editions using zero-shot prompts, and ChatGPT-4 excelled in math questions.

Joshi et al.~\cite{joshi} examined the use of ChatGPT-4 as an educational tool among undergraduate Computer Science students. They found significant inaccuracies in various question types, highlighting potential risks to learning and academic integrity. Despite these challenges, they provided recommendations for students and educators to use ChatGPT-4 constructively to improve education. 
In our work, we evaluated four LLMs on two POSCOMP editions and, despite specific inaccuracies, the high accuracy rates and explanations offered by the models can guide students in their preparation.

Espejel et al.~\cite{espejel} assessed the reasoning ability of ChatGPT-3.5, ChatGPT-4, and BARD in natural language processing tasks. Results showed that ChatGPT-4 outperformed both ChatGPT-3.5 and BARD in zero-shot prompt scenarios for nearly all tasks. Despite its strengths, all three models showed limited proficiency in inductive reasoning and mathematical tasks. 
In our study, we assessed four LLMs on two POSCOMP editions using zero-shot prompts. ChatGPT-4 also performed well in math.

Toyama et al.~\cite{toyama} evaluated ChatGPT-3.5, ChatGPT-4, and Google Bard on 103 questions from the Japan Radiology Board Examination. ChatGPT-4 answered 65\%{} of the questions correctly, significantly outperforming ChatGPT-3.5 (40.8\%{}) and Google Bard (38.8\%{}). ChatGPT-4 excelled in categories requiring lower-order thinking and complex clinical radiology questions. In our work, we evaluated four LLMs on two POSCOMP editions using zero-shot prompts. 
In our evaluation, ChatGPT-4 performed better than Gemini, the successor of Bard.

Nunes et al.~\cite{nunes} evaluated ChatGPT-3.5 and ChatGPT-4 on the National High School Exam (ENEM). Analyzing questions from 2009-2017 and 2022, the study used multiple prompt strategies, including Chain-of-Thought (CoT)~\cite{chain-of-thought}. ChatGPT-4 with CoT achieved 87\%{} accuracy on the 2022 exam, outperforming ChatGPT-3.5 by 11 points. In our study, we evaluated four LLMs on two POSCOMP editions using zero-shot prompts. 
In our work, ChatGPT-4 achieved the highest performance, surpassing all students who took the POSCOMP 2023 exam.

\revisao{
Mendon\c{c}a~\cite{DBLP:journals/toce/Mendonca24} investigated ChatGPT-4 Vision's performance on the 2021 ENADE exam for Computer Science undergraduates in Brazil, showing that the model outperformed the average participant and ranked in the top 10\%, despite limitations in interpreting visual content and complex reasoning.
Our study evaluates multiple state-of-the-art LLMs from OpenAI, Anthropic, and Google on three editions of the POSCOMP exam (2022--2024). We evaluate not only the models' vision capabilities, but also their performance on textual and PDF-based exam content.
Both works demonstrate that advanced LLMs can rival or exceed human performance on Brazilian computer science assessments. Similar patterns emerge, such as strong performance on textual questions, consistent challenges with image-based prompts, and close alignment with human scoring distributions. These complementary results underscore the potential of LLMs as tools for education and assessment analysis, while reinforcing the importance of human oversight in high-stakes academic evaluations.
}

\section{POSCOMP}
\label{sec:poscomp}

The National Exam for Admission to Graduate Studies in Computing (POSCOMP)~\cite{poscomp} is a Brazilian examination designed to assess the foundational competencies of candidates seeking admission to graduate programs in Computer Science (CS) and related fields. First administered in 2000 and organized by the Brazilian Computer Society (SBC) since 2002, POSCOMP plays a crucial role in streamlining the selection process for most graduate computer science programs across the country. The exam is conducted annually, with the \revisao{2024 edition being the 21$^{st}$} organized by SBC (the exam was not held in 2020 and 2021 due to the COVID-19 pandemic).

The exam evaluates three core areas: Mathematics (Figure~\ref{fig:poscomp-12}), Computer Science Fundamentals (Figure~\ref{fig:poscomp-32}), and Computing Technology (Figure~\ref{fig:poscomp-65}). Candidates are required to answer 70 carefully curated multiple-choice questions, aligning with standard curricula of top undergraduate computer science programs in Brazil. The exam lasts four hours. \revisao{The 2023 and 2024 editions of the exam were conducted entirely online, with 617 and 778 participants, respectively.} The previous edition in 2022 was held in person.

\begin{figure}[]
    \centering
    \fcolorbox{black}{gray!20}{
        \parbox{0.45\textwidth}{
            \textbf{QUESTÃO 12} - Determine a distância aproximada entre o ponto J(3, 1) e a reta \(s: 6x - 2y + 11 = 0\).

        A) 1, 3 \\
        B) 2, 6 \\
        C) 4, 3 \\
        D) 12, 1 \\
        E) 18, 5
        }
    }
    \caption{Question 12 from POSCOMP 2023 (Mathematics). Find the approximate distance between a point and a line.}
    \label{fig:poscomp-12}
\end{figure}

\begin{figure}[]
    \centering
    \fcolorbox{black}{gray!20}{
        \parbox{0.45\textwidth}{
            \textbf{QUESTÃO 32} - Um grafo não direcionado no qual todos os pares de vértices são adjacentes, isto é, possui arestas ligando todos os vértices entre si, é um grafo:

            A) Desconexo. \\
            B) Completo. \\
            C) Ponderado. \\
            D) Livre. \\
            E) Hipergrafo.
        }
    }
    \caption{Question 32 from POSCOMP 2022 (Computer Science Fundamentals). An undirected graph in which every pair of vertices is adjacent, meaning there are edges connecting all vertices, is a graph:}
    \label{fig:poscomp-32}
\end{figure}

\begin{figure}[]
    \centering
    \fcolorbox{black}{gray!20}{
        \parbox{0.45\textwidth}{
            \textbf{QUESTÃO 65} - Uma rede conectada à Internet possui a máscara de sub-rede 255.255.255.128. Qual o número máximo de computadores que a rede suporta? \\
            A) 126 \\
            B) 128 \\
            C) 254 \\
            D) 255.255.255.128 \\
            E) 256
        }
    }
    \caption{Question 65 from POSCOMP 2023 (Computing Technology). A network connected to the Internet has a subnet mask of 255.255.255.128. What is the maximum number of computers the network can support?}
    \label{fig:poscomp-65}
\end{figure}

POSCOMP provides participants with detailed individual results, including correct/incorrect answers, overall averages, and standard deviations. Furthermore, official exams and answer keys are published online~\cite{poscomp}. Since 2006, POSCOMP's reach has extended beyond Brazil through a strategic partnership with the Peruvian Computer Society, enabling the exam to be administered in Peru as well. This collaboration broadens opportunities for prospective computer science graduate students.

\section{Evaluation}
\label{sec:evaluation}

In this section, we present the evaluation for detecting solving POSCOMP questions.

\revisao{\subsection{Research Questions}}

The objective of this study is to assess the effectiveness of specific Large Language Models -- ChatGPT-4, Gemini 1.0 Advanced, Claude 3 Sonnet, and Le Chat Mistral -- in solving questions from the POSCOMP 2022 and 2023 exams. The aim is to identify the domains and question types in which these models excel or show deficiencies. To achieve this, the following research questions (RQs) will be addressed:
\begin{itemize}%
\item[\textbf{RQ$_{1}$}] To what extent can ChatGPT-4 solve POSCOMP problems?
\item[\textbf{RQ$_{2}$}] To what extent can Gemini 1.0 Advanced solve POSCOMP problems?
\item[\textbf{RQ$_{3}$}] To what extent can Claude 3 Sonnet solve POSCOMP problems?
\item[\textbf{RQ$_{4}$}] To what extent can Le Chat Mistral solve POSCOMP problems?
\revisao{
\item[\textbf{RQ$_{5}$}] To what extent do recent LLMs match or surpass human performance in POSCOMP exams?
}
\end{itemize}
Correct and incorrect answers provided by each LLM will be counted according to the official answer key provided by POSCOMP~\cite{poscomp}.

\subsection{Planning}
\label{sec:planning}

We evaluated four LLMs: ChatGPT-4, Gemini 1.0 Advanced, Claude 3 Sonnet, and Le Chat Mistral 1.0. In addition to the 2023 POSCOMP exam, these models were also tested with the 2022 exam. 
\revisao{Each exam contains 70 questions. However, one question from each exam was officially invalidated (Question 40 from the 2022 exam and Question 17 from the 2023 exam). These questions were excluded from our evaluation.}
In this evaluation, the LLMs were provided with English translations of the question text and answer options, transcribed via Google Lens and translated using DeepL. The decision to use English translations was driven by the idea that LLMs are often trained on larger datasets in English, which could potentially improve their performance~\cite{intrator}.

Supporting images (class diagrams, circuits, and automata) were included when necessary. Specifically, in the 2022 exam, these were needed for Questions 27, 31, 40, and 43; in the 2023 exam, for Questions 17, 31, 32, and 34. \revisao{Question 40 from the 2022 exam and Question 17 from the 2023 exam were officially invalidated.} Due to its inability to process images, Le Chat Mistral 1.0 was only provided with the translated text. \revisao{The evaluation was conducted between March 16 and March 22, 2024, using a zero-shot prompt approach, where no prior examples were given to the LLM~\cite{prompt-techniques, prompts}. We used the English version of each question as the prompt input. Each exam contains three images: three class diagrams and three circuit diagrams. The only textual information provided in these images includes class names, relationships, methods, and attributes. However, we did not translate the images; instead, we used the original ones from the Portuguese exam. In the textual descriptions, we also encountered limitations when converting certain mathematical expressions and special characters into text prompts, as they appear in some of the questions.}

\subsection{Results}

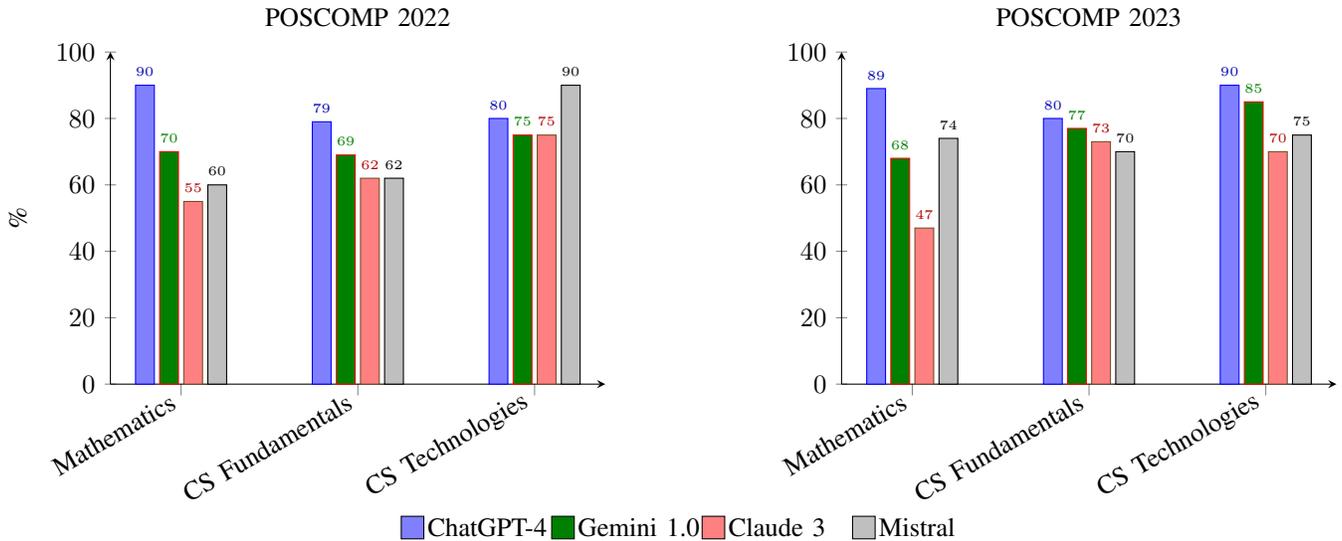
\begin{figure*}[ht]
\centering

\begin{minipage}{0.45\textwidth}
\centering
\begin{tikzpicture}
\begin{axis}[
    axis lines=left,
    ybar,
    bar width=.25cm,
    width=\linewidth,
    height=6cm,
    enlarge x limits=0.2,
    ylabel={\%},
    symbolic x coords={Mathematics, CS Fundamentals, CS Technologies},
    xtick=data,
    ymin=0, ymax=100,
    title={POSCOMP 2022},
    xticklabel style={rotate=30, anchor=east}
]
\addplot+[
    fill=blue!50,
    nodes near coords,
    every node near coord/.append style={font=\tiny, color=blue!70!black}
] coordinates {(Mathematics, 90.0) (CS Fundamentals, 79) (CS Technologies, 80.0)};
\addplot+[
    fill=green!50!black,
    nodes near coords,
    every node near coord/.append style={font=\tiny, color=green!50!black}
] coordinates {(Mathematics, 70.0) (CS Fundamentals, 69.0) (CS Technologies, 75.0)};
\addplot+[
    fill=red!50,
    nodes near coords,
    every node near coord/.append style={font=\tiny, color=red!70!black}
] coordinates {(Mathematics, 55.0) (CS Fundamentals, 62) (CS Technologies, 75.0)};
\addplot+[
    fill=gray!50,
    nodes near coords,
    every node near coord/.append style={font=\tiny, color=black}
] coordinates {(Mathematics, 60.0) (CS Fundamentals, 62) (CS Technologies, 90.0)};
\end{axis}
\end{tikzpicture}
\end{minipage}%
\hfill
\begin{minipage}{0.45\textwidth}
\centering
\begin{tikzpicture}
\begin{axis}[
    axis lines=left,
    ybar,
    bar width=.25cm,
    width=\linewidth,
    height=6cm,
    enlarge x limits=0.2,
    symbolic x coords={Mathematics, CS Fundamentals, CS Technologies},
    xtick=data,
    ymin=0, ymax=100,
    title={POSCOMP 2023},
    xticklabel style={rotate=30, anchor=east}
]
\addplot+[
    fill=blue!50,
    nodes near coords,
    every node near coord/.append style={font=\tiny, color=blue!70!black}
] coordinates {(Mathematics, 89) (CS Fundamentals, 80.0) (CS Technologies, 90.0)};
\addplot+[
    fill=green!50!black,
    nodes near coords,
    every node near coord/.append style={font=\tiny, color=green!50!black}
] coordinates {(Mathematics, 68) (CS Fundamentals, 77) (CS Technologies, 85.0)};
\addplot+[
    fill=red!50,
    nodes near coords,
    every node near coord/.append style={font=\tiny, color=red!70!black}
] coordinates {(Mathematics, 47) (CS Fundamentals, 73) (CS Technologies, 70.0)};
\addplot+[
    fill=gray!50,
    nodes near coords,
    every node near coord/.append style={font=\tiny, color=black}
] coordinates {(Mathematics, 74) (CS Fundamentals, 70.0) (CS Technologies, 75.0)};
\end{axis}
\end{tikzpicture}
\end{minipage}

\vspace{-0.2cm}
\begin{center}
\parbox{0.9\textwidth}{
\centering
\begin{tikzpicture}[baseline, inner sep=0pt, outer sep=0pt]
\filldraw[fill=blue!50, draw=black] (0,0) rectangle (0.3,0.3); \node[anchor=west] at (0.35,0.15) {ChatGPT-4};
\filldraw[fill=green!50!black, draw=black] (2,0) rectangle (2.3,0.3); \node[anchor=west] at (2.35,0.15) {Gemini 1.0};
\filldraw[fill=red!50, draw=black] (4,0) rectangle (4.3,0.3); \node[anchor=west] at (4.35,0.15) {Claude 3};
\filldraw[fill=gray!50, draw=black] (6,0) rectangle (6.3,0.3); \node[anchor=west] at (6.35,0.15) {Mistral};
\end{tikzpicture}
}
\end{center}

\vspace{-0.4cm}
\caption{Comparison of Old LLMs' performance on POSCOMP 2022 and 2023 by topic (in percentage).}
\label{fig:old-llms-performance}
\end{figure*}

Next we present the evaluation results. For the 2022 exam (Figure~\ref{fig:old-llms-performance}), ChatGPT-4 stands out as the leader in correct answers, achieving a total score of 57 correct responses out of \revisao{69} questions. This performance is driven by its scores in Mathematics (18 out of 20) and Computer Science (CS) Fundamentals (23 out of 30). Gemini 1.0 Advanced had the second-highest number of correct answers, scoring 49 out of \revisao{69}, trailing behind ChatGPT-4 across all areas. Le Chat Mistral and Claude 3 Sonnet had comparable results to Gemini, with 48 and 44 correct answers, respectively. However, Le Chat Mistral excels in Computing Technology, scoring 18 out of 20, surpassing all other models in this domain.

For the 2023 exam, presented in Figure~\ref{fig:old-llms-performance}, ChatGPT-4 once again achieved the highest score with a total of 59 correct answers out of \revisao{69} questions. Gemini consistently demonstrated strong performance across all areas, achieving a total score of 53 out of \revisao{69}. Le Chat Mistral reached 50 correct answers, while Claude 3 Sonnet lagged slightly behind with 45 correct answers.

\subsection{Discussion}

In this section, we analyze the results achieved.

\subsubsection{Correctness}

In the 2022 exam, a total of 31 questions received the same correct answer from all models, while in the 2023 exam, this increased to 36 questions. This indicates a slight improvement in the collective performance of the models from one year to the next. However, some questions remained challenging, as evidenced by five questions in 2022 and six in 2023 that none of the models managed to solve.

As shown in Table~\ref{tab:unique-questions-correct}, ChatGPT-4 answered six questions correctly by itself, which no other LLM answered, in the 2022 exam and repeated this achievement in the 2023 exam. Similarly, Gemini successfully answered two questions on its own in 2022 and two more in 2023. Claude answered one question alone in the 2022 exam.

\begin{table}[]
\centering
\caption{Number of questions answered correctly by only one model.}
\label{tab:unique-questions-correct}

\begin{tabular}{|c|r|r|}
\hline
\rowcolor[HTML]{333333} 
{\color[HTML]{FFFFFF} \textbf{LLM}} & \multicolumn{1}{c|}{\cellcolor[HTML]{333333}{\color[HTML]{FFFFFF} \textbf{2022}}} & \multicolumn{1}{c|}{\cellcolor[HTML]{333333}{\color[HTML]{FFFFFF} \textbf{2023}}} \\ \hline
ChatGPT                             & 6                                                                                 & 6                                                                                 \\ \hline
Gemini                              & 2                                                                                 & 2                                                                                 \\ \hline
Claude                              & 1                                                                                 & 0                                                                                 \\ \hline
Mistral                             & 0                                                                                 & 0                                                                                 \\ \hline
\end{tabular}

\end{table}

\begin{table}[]
\centering
\caption{Number of questions where models selected multiple answers as correct.}
\label{tab:multiple-answers}

\begin{tabular}{|c|r|r|}
\hline
\rowcolor[HTML]{333333} 
{\color[HTML]{FFFFFF} \textbf{LLM}} & \multicolumn{1}{c|}{\cellcolor[HTML]{333333}{\color[HTML]{FFFFFF} \textbf{2022}}} & \multicolumn{1}{c|}{\cellcolor[HTML]{333333}{\color[HTML]{FFFFFF} \textbf{2023}}} \\ \hline
ChatGPT                             & 0                                                                                 & 0                                                                                 \\ \hline
Gemini                              & 1                                                                                 & 3                                                                                 \\ \hline
Claude                              & 3                                                                                 & 5                                                                                 \\ \hline
Mistral                             & 1                                                                                 & 5                                                                                 \\ \hline
\end{tabular}

\end{table}

The models sometimes provided multiple answers for a question (Table~\ref{tab:multiple-answers}) or failed to provide any answer (Table~\ref{tab:no-answer}). For instance, Claude often selected multiple answers in both tests, indicating a tendency to be less decisive or perhaps more exploratory in its response strategy. In contrast, Mistral, ChatGPT-4, and Gemini tended to leave some questions unanswered in 2022, suggesting possible gaps in their knowledge or caution in their response approach.

\begin{table}[]
\centering
\caption{Number of questions where the models did not select a correct option.}
\label{tab:no-answer}

\begin{tabular}{|c|r|r|}
\hline
\rowcolor[HTML]{333333} 
{\color[HTML]{FFFFFF} \textbf{LLM}} & \multicolumn{1}{c|}{\cellcolor[HTML]{333333}{\color[HTML]{FFFFFF} \textbf{2022}}} & \multicolumn{1}{c|}{\cellcolor[HTML]{333333}{\color[HTML]{FFFFFF} \textbf{2023}}} \\ \hline
ChatGPT                             & 5                                                                                 & 1                                                                                 \\ \hline
Gemini                              & 4                                                                                 & 2                                                                                 \\ \hline
Claude                              & 1                                                                                 & 1                                                                                 \\ \hline
Mistral                             & 6                                                                                 & 3                                                                                 \\ \hline
\end{tabular}

\end{table}

\begin{figure*}[ht]
\begin{center}
\leavevmode
\scalebox{0.5}{
\includegraphics{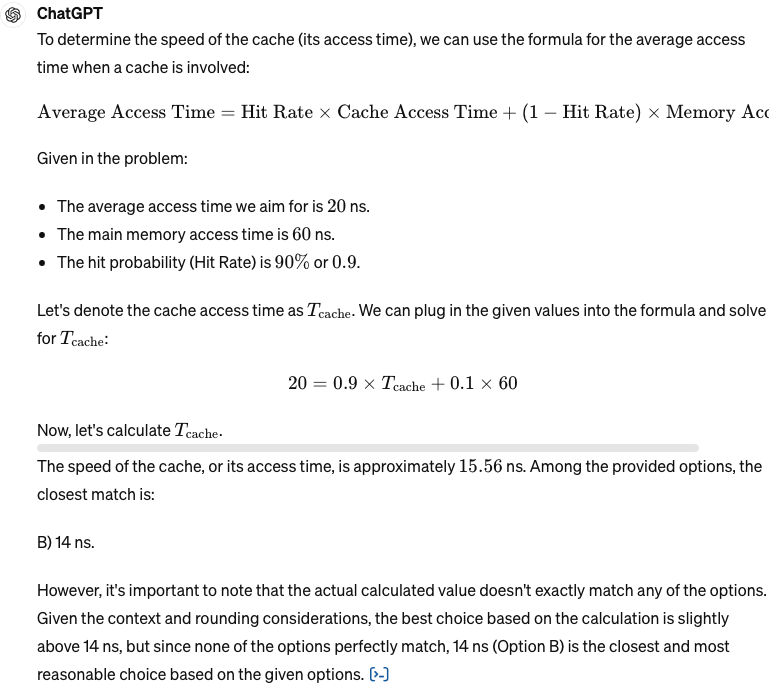}}
\caption{ChatGPT-4 answer to Question 29 from POSCOMP 2023.}
\label{fig:poscomp-29-2023-chatgpt-answer}
\end{center}
\end{figure*}

The question shown in Figure~\ref{fig:poscomp-29-2023} (Question 29 from POSCOMP 2023) was not correctly answered by any of the LLMs evaluated in our study. The goal is to find the cache speed (access time) if we expect a 90\%{} hit rate. All LLMs selected option B, but according to POSCOMP, the correct answer is C. ChatGPT-4's response is displayed in Figure~\ref{fig:poscomp-29-2023-chatgpt-answer}. All four models calculated a result of \texttt{15.56} ns. ChatGPT-4 and Mistral indicate that they chose the nearest option. Gemini and Claude, however, do not provide an explanation to the user for why they selected an answer that differs from the calculated value. \revisao{In scenarios like this, we recommend the problem setters could reavaluate it to improve its statement and confirm the correct answer.}

\begin{figure}[]
    \centering
    \fcolorbox{black}{gray!20}{
        \parbox{0.45\textwidth}{
            \textbf{QUESTÃO 29} - O computador tem um tempo de acesso à memória principal de 60 ns. Queremos reduzir esse tempo para 20 ns adicionando no cache. Qual a velocidade do cache (tempo de acesso) se pudermos esperar uma probabilidade de 90\%{} de acerto? \\
            A) 04 ns \\
            B) 14 ns \\
            C) 24 ns \\
            D) 54 ns \\
            E) 84 ns
        }
    }
    \caption{Question 29 from POSCOMP 2023. The computer has a main memory access time of 60 ns. We want to reduce this time to 20 ns by adding a cache. What should be the cache speed (access time) if we expect a 90\%{} hit rate?}
    \label{fig:poscomp-29-2023}
\end{figure}

\begin{figure*}[ht]
\begin{center}
\leavevmode
\scalebox{0.6}{

\includegraphics{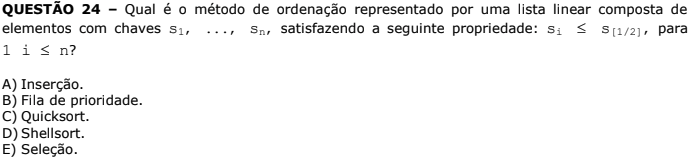}

}
\caption{Question 24 from POSCOMP 2022. What is the sorting method represented by a linear list consisting of elements with keys \( s_1, \ldots, s_n \), satisfying the following property: \( s_i \leq s_{\lceil \revisao{1}/2 \rceil} \), for \( 1 i \leq n \)?
}
\label{fig:poscomp-24-2022-gpt}
\end{center}
\end{figure*}

Claude correctly answered Question 24 from POSCOMP 2022, which none of the other LLMs answered correctly. The question is presented in Figure~\ref{fig:poscomp-24-2022-gpt}. The goal is to find the sorting algorithm. Notice the relationship between \texttt{i} and \texttt{1} is unclear. It appears that some information is missing in this question. \revisao{Moreover, it uses 1/2 as an index instead of the variable \textit{i}.} Claude's answer asserts that \texttt{i} is greater than or equal to \texttt{1}. However, Gemini identified option E as the correct choice, while ChatGPT-4 and Mistral stated that none of the algorithms possessed the specific property. Consequently, they were unable to select an accurate answer from the available options. We included this property (\texttt{1$\leq{}$i}) by adding the missing operator, and asked ChatGPT-4 again. It then correctly chose option B. \revisao{In scenarios like this, where the majority of models answer incorrectly, problem setters may consider revising the question statements based on the models' responses.}

\begin{figure*}[ht]
\begin{center}
\leavevmode
\scalebox{0.6}{
\includegraphics{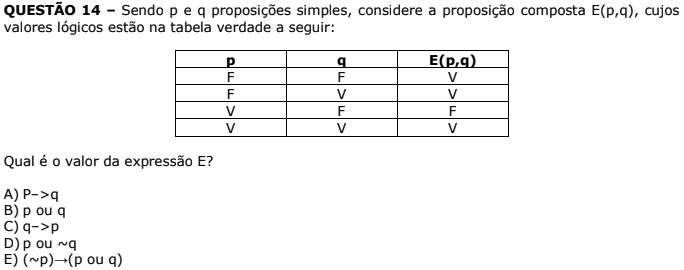}

}
\caption{Question 14 from POSCOMP 2023. Given that p and q are simple propositions, consider the compound proposition E(p,q), whose logical values are in the following truth table. What is the value of the expression E?
}
\label{fig:poscomp-14-2023-gpt}
\end{center}
\end{figure*}

ChatGPT-4 correctly answered Question 14 from POSCOMP 2023 (see Figure~\ref{fig:poscomp-14-2023-gpt}), which none of the other LLMs answered accurately. The goal is to find an expression from the thruth table. ChatGPT-4 constructed the truth table for all options and explained why option A was correct.
Gemini asserted that options A, B, and E all matched the pattern and proceeded to further testing to identify the exact equivalent expression. It eventually concluded that none of the available options perfectly represented the truth table, suggesting that there might be an error in the question or that a different logical expression type was needed to represent the table.
Mistral chose option B as correct but offered a very brief explanation, failing to clarify the reasoning process. Additionally, Gemini and Mistral did not construct a truth table in their explanations.

\begin{figure*}[ht]
\begin{center}
\leavevmode
\scalebox{0.55}{
\includegraphics{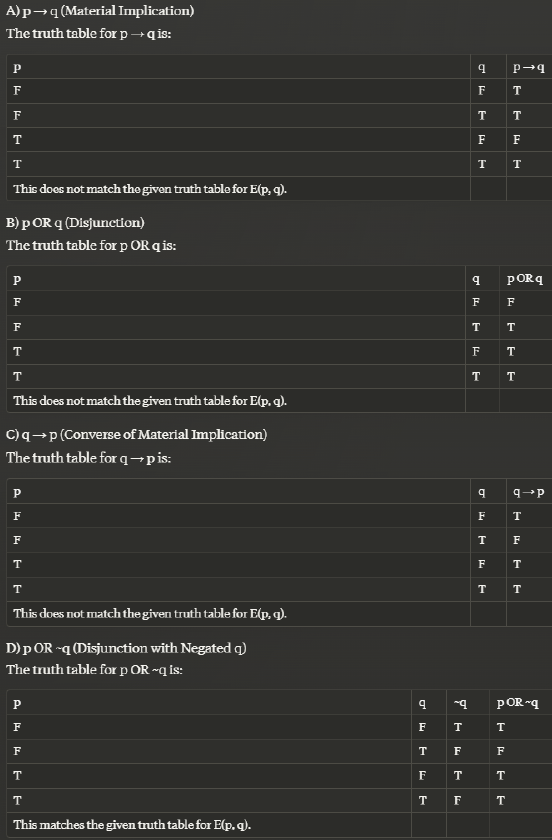}

}
\caption{Claude answer to Question 14 from POSCOMP 2023.}
\label{fig:poscomp-14-2023-gpt-claude}
\end{center}
\end{figure*}

Claude constructed a truth table for all options and accurately produced option A's truth table. It correctly interprets that \texttt{V} in Portuguese stands for `true'. However, it mistakenly concluded that this option did not match the one presented in the question, and it incorrectly selected option D (see Figure~\ref{fig:poscomp-14-2023-gpt-claude}).

\revisao{
Across both POSCOMP exams, there are six valid questions that include images. Despite lacking image support, Le Chat Mistral 1.0 correctly answered 3 out of the 6 image-based questions. In one of them, which involved a UML class diagram representing the concepts of chair, sofa, rack, and table, Mistral successfully identified the correct alternative without accessing the diagram itself. Based solely on the answer options and its prior knowledge, it correctly inferred that the option stating that the class \textit{Chair} inherits from the class \textit{Table} was incorrect. A similar question from the 2022 exam featured a class diagram involving books and chapters; once again, Mistral correctly selected the answer by analyzing the alternatives alone. In the 2023 POSCOMP, a question related to digital circuits provided additional textual information that explained the associated image, which helped Mistral arrive at the correct response.
Claude 3 also answered 3 out of the 6 image-based questions correctly across both exams. Gemini 1.0 answered two questions correctly, while ChatGPT-4 achieved the highest performance, correctly answering 5 out of the 6 questions involving images.
}

\subsubsection{Explanation}

Next, we explore how LLMs can enhance the explanation of each response. The data presented in Tables~\ref{tab:explanation-2022}~and~\ref{tab:explanation-2023} show the number of questions from POSCOMP 2022 and 2023, respectively, using the following criteria based on their answers:
\begin{itemize}
    \item \textit{Explains the Topic:} Indicates that the LLM's response contains some explanation of the question's topic.
    \item \textit{Explains the Selected Option:} Indicates that the LLM's response provides an explanation for the option chosen as correct by the model.
    \item \textit{Explains Each Option:} Indicates that the LLM's response contains explanations for the other options that weren't chosen as correct (e.g., why they are incorrect).
    \item \textit{Only the Option:} Indicates that the model only wrote the correct answer in its response, without providing additional explanation.
\end{itemize}

\begin{table}[]
\centering
\caption{Models' responses to POSCOMP 2022 by level of explanation.}
\label{tab:explanation-2022}

\begin{tabular}{|c|c|c|c|c|}
\hline
\rowcolor[HTML]{333333} 
{\color[HTML]{FFFFFF} \textbf{Explanation}} & {\color[HTML]{FFFFFF} \textbf{ChatGPT}} & {\color[HTML]{FFFFFF} \textbf{Gemini}} & {\color[HTML]{FFFFFF} \textbf{Claude}} & {\color[HTML]{FFFFFF} \textbf{Mistral}} \\ \hline
Topic                                       & 38/69                                   & 52/69                                  & 50/69                                  & 38/69                                   \\ \hline
Selec. option                               & 22/69                                   & 48/69                                  & 68/69                                  & 63/69                                   \\ \hline
Each option                                 & 12/69                                   & 29/69                                  & 19/69                                  & 5/69                                    \\ \hline
Only option                                 & 10/69                                   & 0/69                                   & 0/69                                   & 2/69                                    \\ \hline
\end{tabular}

\end{table}

In 2022, Gemini stood out for frequently explaining both the topic (52/69) and the correct option (48/69), significantly more often than ChatGPT-4, Claude, and Mistral in similar categories. Gemini and Claude did not respond with only the correct option, suggesting a more in-depth approach to their answers. Claude, notably strong in explaining the correct option (68/69), indicates a focused clarity in its responses. In contrast, Mistral and ChatGPT-4 occasionally opted for less detailed answers, with ChatGPT-4 providing only the correct option in 10 out of 69 cases and Mistral in 2 out of 69 cases.

\begin{table}[]
\centering
\caption{Models' responses to POSCOMP 2023 by level of explanation.}
\label{tab:explanation-2023}

\begin{tabular}{|c|c|c|c|c|}
\hline
\rowcolor[HTML]{333333} 
{\color[HTML]{FFFFFF} \textbf{Explanation}} & {\color[HTML]{FFFFFF} \textbf{ChatGPT}} & {\color[HTML]{FFFFFF} \textbf{Gemini}} & {\color[HTML]{FFFFFF} \textbf{Claude}} & {\color[HTML]{FFFFFF} \textbf{Mistral}} \\ \hline
Topic                                       & 49/69                                   & 52/69                                  & 63/69                                  & 34/69                                   \\ \hline
Selec. option                               & 30/69                                   & 31/69                                  & 63/69                                  & 58/69                                   \\ \hline
Each option                                 & 19/69                                   & 32/69                                  & 25/69                                  & 12/69                                   \\ \hline
Only option                                 & 3/69                                    & 0/69                                   & 0/69                                   & 5/69                                    \\ \hline
\end{tabular}

\end{table}

In the 2023 exam, Claude exhibited consistency, particularly in explaining both the topic (63/69) and the correct option (63/69). This consistency highlights Claude's reliable and comprehensive output in handling queries. Gemini maintained strong performance in explaining the topic but showed a slight decline in detailing the correct option compared to 2022. ChatGPT-4 demonstrated improvement in topic explanation, increasing from 38/69 to 49/69, indicating enhanced contextual information. Mistral, however, showed modest overall performance, with a focus on explaining the correct option.

These trends highlight the different strengths and strategies employed by the models. While models like Claude and Gemini consistently aim for comprehensive explanations, others like Mistral and ChatGPT-4 may focus more on direct responses under certain conditions. This variation in approach may reflect the underlying design and intended application of each model, catering to different user needs for explanation and detail.

\subsubsection{Topics}

In the 2022 exam, ChatGPT-4 maintained a high success rate but was closely followed by other models. ChatGPT-4 achieved 18/20 in Mathematics, 23/\revisao{29} in Computer Science Fundamentals, and 16/20 in Computing Technology, resulting in a total score of 57/69. Gemini, Claude, and Mistral also performed well in this test, with scores not far behind ChatGPT-4: 49/69, 44/69, and 48/69, respectively.
In the 2023 exam, ChatGPT-4 once again led with 59/69, demonstrating its versatility and consistency across general topics. However, Gemini, Claude, and Mistral also performed well, scoring 53/69, 45/69, and 50/69, respectively.

In the 2022 exam, there was more parity among models in specific topics. While ChatGPT-4 excelled in many areas, Gemini, Claude, and Mistral also achieved high success rates. For example, in \revisao{8 questions related to} Analytical Geometry, Differential and Integral Calculus, and Artificial Intelligence, all models performed strongly with success rates above 60\%{}. However, differences emerged in \revisao{6 questions related to} specific areas like Discrete Mathematics and Formal Languages, Automata, and Computability, where ChatGPT-4 outperformed the other models.
In the 2023 exam, ChatGPT-4 maintained its lead across several specific topics, achieving 100\% accuracy in \revisao{10 questions related to} Linear Algebra, Analytical Geometry, and Mathematical Logic. Gemini and the other models showed mixed performance, but achieved high accuracy in selected areas, \revisao{particularly in 8 questions related to} Combinatorial Analysis, Probability and Statistics, and Graph Theory.

In Mathematics overall, Claude had a weaker performance compared to the other models, with a 50\%{} success rate in both exams. This lower performance may be due to factors like a lack of training in specific mathematical concepts or less emphasis on rigorous mathematical problem-solving in its training data.
Overall, these results suggest that ChatGPT-4 generally outperforms other models across general topics. However, success rates in specific areas vary, indicating that different models excel in distinct domains.

\subsubsection{Metamorphic Testing}
\label{sec:metamorphic}

\revisao{
A key threat to validity when evaluating foundation models is data contamination~\cite{threats-llms-icse-nier-2024}, which occurs when a model has been exposed to identical or similar examples during training, potentially leading to inflated performance results. To mitigate this risk, we adopt metamorphic testing~\cite{metamorphic-testing,metamorphic-testing-2} to assess the robustness and reliability of the models. This approach involves generating new data samples by applying controlled metamorphic transformations to the original validation or test questions. These transformations preserve the semantic meaning of the original question while introducing syntactic variations. The primary goal is to evaluate the model’s resilience and its ability to consistently identify the correct answer, even when the question is presented in a structurally altered but semantically equivalent form.
}

For this experiment, a sample of 10 questions was selected: five from POSCOMP 2022 and five from POSCOMP 2023. From the 2022 exam, Questions 10, 20, 22 (shown in its original version in Figure~\ref{fig:poscomp-22-original} and modified in Figure~\ref{fig:poscomp-22-metamorphic}), 30, and 57 were selected. From the 2023 exam, questions 5, 19, 21, 39, and 65 were chosen. The 2022 exam questions were altered as follows:
\begin{itemize}
    \item Question 10: The expression ((2x+3)$^{3}$(3x-2)$^{2}$) was changed to ((2x+3)$^{6}$(3x-2)$^{2}$), and the expression x$^{5}$ +5 was changed to x$^{8}$+5. The options were modified from A) 72, B) 19, C) 9, D) 8, E) 0 to A) 1024, B) 720, C) 576, D) 12, E) 0.
    \item Question 20: The values of t were changed from 2, 3, 4, 5, 6, 7 to 3, 4, 5, 6, 7, 8. The options were changed from A) 4.5 s, B) 5.0 s, C) 1.0 s, D) 0.9 s, E) 5.4 s to A) 0.9 s, B) 5.4 s, C) 1.0 s, D) 6.4 s, E) 5.0 s.
    \item Question 22: In the statement, f3(n) = O(2$^{\wedge}$n) was changed to f3(n) = O(2$^{\wedge}$(3$^{\wedge}$n)). Options B and C were swapped with options D and E.
    \item Question 30: Statement III was modified from ``Integer types are used to store values that belong to the set of natural numbers (without fractional parts)'' to ``The boolean type stores only the `False' value and fractional numbers.'' Options A) Only I, B) Only II, and C) Only III were changed to A) Only II, B) Only III., and C) Only I and II.
    \item Question 57: Statement I was changed from ``Mapping images as textures (surface texture) is a technique that uses a 2D coordinate system.'' to ``Ray tracing is a technique used to generate textures by tracing the path of shadow circles through an image plane.'' Option B) Only III was changed to B) Only II.
\end{itemize}

\begin{figure}[]
    \centering
    \fcolorbox{black}{gray!20}{
        \parbox{0.45\textwidth}{
        \textbf{QUESTÃO 22} - Considere as funções a seguir:

        \text{f1}(n) = O(n) \\
        \text{f2}(n) = O(n!) \\
        \text{f3}(n) = O(2$^{\wedge}$n) \\
        \text{f4}(n) = O(n$^{\wedge}$2) \\

        A ordem dessas funções, por ordem crescente de taxa de crescimento, é:

        A) f2 - f1 - f3 - f4. \\
        B) f3 - f2 - f4 - f1. \\
        C) f1 - f4 - f3 - f2. \\
        D) f1 - f4 - f2 - f3. \\
        E) f4 - f3 - f1 - f2. \\
            
        }
    }
    \caption{Question 22 from POSCOMP 2022 in its original version. Consider the following functions. Arrange them in ascending order according to their growth rate.}
    \label{fig:poscomp-22-original}
\end{figure}

For the 2023 exam, the questions were modified as follows:
\begin{itemize}
    \item Question 5: The number of regions was changed from 10 to 8. Options were altered from A) 1024, B) 10, C) 100, D) 512, and E) 20 to A) 80, B) 64, C) 8, D) 1024, and E) 256.
    \item Question 19: The number of families was changed from 5, 6, 8, 4, and 2 to 65, 43, 9, 73, and 12. Options A) 1.12, C) 2.11, and E) 3.21 were changed to A) 1.62, C) 2.71, and E) 3.02.
    \item Question 21: Option A was swapped with option C, option D was swapped with option B, and option B was swapped with option E. A new option D was added: ``If the time required by an algorithm on all inputs of size n is, at most, 5n$^{3}$ + 3n, the asymptotic complexity is O(n$^{3}$).''
    \item Question 39: Statements I and III were swapped. In statement II, the word ``efficient'' was changed to ``INEFFICIENT.''
    \item Question 65: The subnet mask in the question statement was changed from 255.255.255.128 to 255.255.255.64. Options A) 126, B) 128, D) 255.255.255.128, and E) 256 were changed to A) 128, B) 62, D) 255.255.255.64, and E) 255.255.255.128.
\end{itemize}

\begin{figure}[]
    \centering
    \fcolorbox{black}{gray!20}{
        \parbox{0.45\textwidth}{

        \textbf{QUESTÃO 22} - Considere as funções a seguir:

        \text{f1}(n) = O(n) \\
        \text{f2}(n) = O(n!) \\
        \text{f3}(n) = O(2$^{\wedge}${(3$^{\wedge}$n)}) \\
        \text{f4}(n) = O(n$^{\wedge}$2) \\

        A ordem dessas funções, por ordem crescente de taxa de crescimento, é::

        A) f2 - f1 - f3 - f4. \\
        B) f1 - f4 - f2 - f3. \\
        C) f4 - f3 - f1 - f2. \\
        D) f3 - f2 - f4 - f1. \\
        E) f1 - f4 - f3 - f2.
            
        }
    }
    \caption{Question 22 from POSCOMP 2022, translated and modified for the metamorphic test. Consider the following functions. Arrange them in ascending order according to their growth rate.}
    \label{fig:poscomp-22-metamorphic}
\end{figure}

To ensure accuracy in the modified responses, the authors analyzed each modified question.
For the 2022 exam, as shown in Table~\ref{tab:metamorphic-testing}, both \revisao{ChatGPT-4 and Gemini 1.0 Advanced} demonstrated flawless performance, aligning perfectly with the correct answers in all five questions. However, Claude and Mistral displayed some discrepancies. Specifically, Claude answered Question 10 incorrectly. Besides getting Question 10 wrong, Mistral also selected a wrong answer on Questions 22 and 57. These errors relate to interpretation issues in the models' responses.

In the transition to the 2023 exam, as shown in Table~\ref{tab:metamorphic-testing}, ChatGPT-4 and Gemini 1.0 Advanced maintained their perfect accuracy from the previous exam, answering all five questions correctly. In contrast, although Claude answered most questions correctly, it failed on Question 21 by selecting both `A' and `D' as correct. Mistral has the same error, indicating a potential flaw in its processing or a misunderstanding of the question requirements.

\begin{table}[]
\centering
\caption{Results of POSCOMP metamorphic tests.}
\label{tab:metamorphic-testing}

\begin{tabular}{|c|c|c|c|c|c|}
\hline
\rowcolor[HTML]{333333} 
{\color[HTML]{FFFFFF} \textbf{Year}} & {\color[HTML]{FFFFFF} \textbf{Qu.}} & {\color[HTML]{FFFFFF} \textbf{ChatGPT}} & {\color[HTML]{FFFFFF} \textbf{Gemini}} & {\color[HTML]{FFFFFF} \textbf{Claude}} & {\color[HTML]{FFFFFF} \textbf{Mistral}} \\ \hline
2022                        & 10                              & \checkmark                            & \checkmark                           & $\times{}$                            & $\times{}$                             \\ \hline
2022                        & 20                              & \checkmark                            & \checkmark                           & \checkmark                           & \checkmark                            \\ \hline
2022                        & 22                              & \checkmark                            & \checkmark                           & \checkmark                           & $\times{}$                             \\ \hline
2022                        & 30                              & \checkmark                            & \checkmark                           & \checkmark                           & \checkmark                            \\ \hline
2022                        & 57                              & \checkmark                            & \checkmark                           & \checkmark                           & $\times{}$                             \\ \hline
2023                        & 5                               & \checkmark                            & \checkmark                           & \checkmark                           & \checkmark                            \\ \hline
2023                        & 19                               & \checkmark                            & \checkmark                           & \checkmark                           & \checkmark                            \\ \hline
2023                        & 21                              & \checkmark                            & \checkmark                           & $\times{}$                            & $\times{}$                             \\ \hline
2023                        & 39                              & \checkmark                            & \checkmark                           & \checkmark                           & \checkmark                            \\ \hline
2023                        & 65                              & \checkmark                            & \checkmark                           & \checkmark                           & \checkmark                            \\ \hline
\end{tabular}

\end{table}

The overall results highlight that ChatGPT-4 and Gemini 1.0 Advanced are robust in their adaptability and accuracy in these exams, maintaining consistent performance across both exams. Claude and Mistral showed less consistent performance, particularly in the 2022 exam and Question 21 in 2023, suggesting that while they generally perform well, they could benefit from refinement in handling ambiguous or complex scenarios.

\subsubsection{\revisao{Visual Reasoning}}

\revisao{We assess the visual reasoning capabilities of LLMs by presenting POSCOMP 2023 questions entirely as images.
In this study, we did not translate the questions into English; instead, we used the original statements in Portuguese. The prompt used was simply: ``Qual é a resposta?''. Each prompt included the image of the corresponding question. The evaluation was conducted between March 16 and March 22, 2024, using a zero-shot prompting approach, where no prior examples were provided to the LLM~\cite{prompt-techniques, prompts}.
For this evaluation, we used the 2023 POSCOMP exam and the following model versions: \revisao{ChatGPT-4 and Gemini 1.0 Advanced}. Each question was captured as an image using the ``Snip \&{} Sketch'' tool on Windows 10. The screenshots included the question statement, answer options, and any associated images or tables, as presented in the original exam. 
}

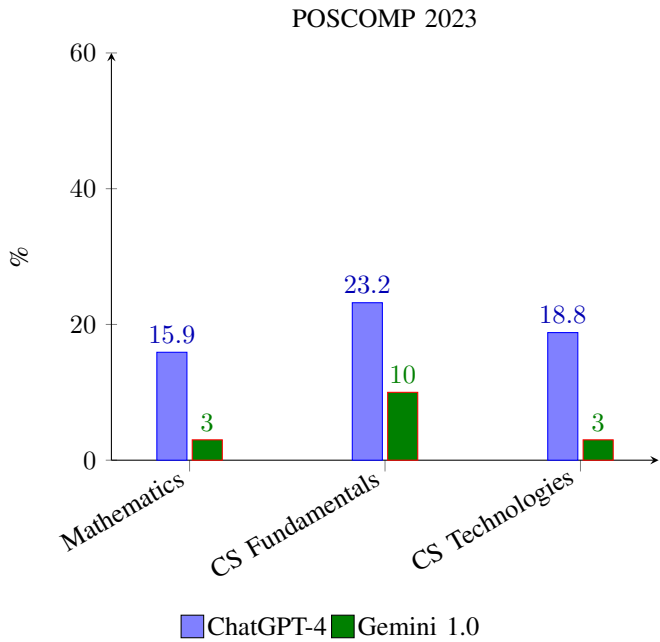
\begin{figure}[ht]
\centering
\begin{tikzpicture}
\begin{axis}[
    axis lines=left,
    ybar,
    bar width=.4cm,
    width=\linewidth,
    height=7cm,
    enlarge x limits=0.2,
    ylabel={\%},
    symbolic x coords={Mathematics, CS Fundamentals, CS Technologies},
    xtick=data,
    ymin=0, ymax=60,
    nodes near coords,
    every node near coord/.append style={font=\normalsize},
    title={POSCOMP 2023},
    xticklabel style={rotate=30, anchor=east},
]
\addplot+[
    fill=blue!50,
    nodes near coords,
    every node near coord/.append style={color=blue!70!black}
] coordinates {(Mathematics, 15.9) (CS Fundamentals, 23.2) (CS Technologies, 18.8)};

\addplot+[
    fill=green!50!black,
    nodes near coords,
    every node near coord/.append style={color=green!50!black}
] coordinates {(Mathematics, 3) (CS Fundamentals, 10) (CS Technologies, 3)};
\end{axis}
\end{tikzpicture}

\vspace{0.2cm}

\begin{tikzpicture}[baseline, inner sep=0pt, outer sep=0pt]
\filldraw[fill=blue!50, draw=black] (0,0) rectangle (0.3,0.3); \node[anchor=west] at (0.35,0.15) {ChatGPT-4};
\filldraw[fill=green!50!black, draw=black] (2,0) rectangle (2.3,0.3); \node[anchor=west] at (2.35,0.15) {Gemini 1.0};
\end{tikzpicture}

\caption{Performance of LLMs with image-based prompts on POSCOMP 2023 by topic (in percentage).}
\label{fig:poscomp-images-2023}
\end{figure}

ChatGPT-4 demonstrated ability to tackle POSCOMP questions, achieving a success rate of approximately 57.9\%{} (40 out of 69 questions). 
It showed proficiency across various subjects, with notable strengths in Computing Technology, Computer Science Fundamentals, and specific areas of Mathematics like Linear Algebra and Probability and Statistics. 
Gemini 1.0 Advanced encountered significant difficulties with the POSCOMP questions, achieving only 11 correct answers out of 69, translating to a success rate of around 15.9\%{}. 
The model performed poorly across all subjects, particularly in Mathematics and Computing Technology. Gemini's challenges were exacerbated by hallucination issues during the test, leading to incorrect or nonsensical answers. 
As observed, both LLMs faced challenges in the evaluation using a prompt with a screenshot of the question statement \revisao{(see Figure~\ref{fig:poscomp-images-2023})}. This is an area where LLMs can improve.

\revisao{
\subsubsection{Latest Models}

Given the rapid advancement of LLMs, we extended our evaluation in April 2025 to include the 2022–2024 POSCOMP exams, leveraging cutting-edge models from OpenAI, Google, and Anthropic. This study includes the most recent POSCOMP exam, which was not covered in our previous evaluation. We followed the same experimental setup described in Section~\ref{sec:planning}. The models were accessed through their respective web interfaces, and evaluations were conducted using default parameters. The LLMs assessed include OpenAI's o1 and o3-mini-high, Google's Gemini 2.5 Pro Experimental, and Anthropic's Claude 3.7 Sonnet.
This evaluation, unlike previous ones, aimed to investigate three key aspects: the ability to analyze PDF documents, the performance on batches of multiple questions, and the handling of content in Portuguese.

Each model was prompted in Portuguese with the instruction: ``Responda a prova em anexo.'' To manage context window limitations and daily usage constraints, each exam (comprising 70 questions) was divided into smaller PDF files -- each with a maximum of four pages and no more than 20 questions. Consequently, each POSCOMP exam, originally written in Portuguese, was divided into five separate PDFs. We submitted these five files to each model, effectively prompting them to answer an entire POSCOMP exam across five interactions. The evaluation was conducted during the last week of March and the first week of April 2025. 

All LLMs provide a single answer per question, unlike earlier versions that returned multiple responses (see Table~\ref{tab:multiple-answers}). In the 2022 exam, only one question was answered correctly exclusively by Claude 3.7 Sonnet, and four were uniquely answered correctly by Gemini 2.5 Pro Experimental. In 2023, Gemini 2.5 Pro Experimental was the only model to correctly answer three questions that none of the other models could. In 2024, o3-mini-high and Claude 3.7 each answered one question correctly that no other model did, while Gemini 2.5 correctly answered three such questions. These cases highlight the complementary strengths of different models and suggest potential benefits in aggregating their responses.

Figure~\ref{fig:poscomp-results} presents a comparative analysis of the four state-of-the-art LLMs on the POSCOMP exams from 2022 to 2024. Results are categorized by topic: Mathematics, Computer Science Fundamentals, and Computer Science Technologies. Each bar reflects the percentage of correct answers achieved by each model for a given topic and year.

Overall, Gemini 2.5 Pro Experimental consistently delivered the best performance across all three years. It improved upon its previous version and outperformed OpenAI's model (see Figure~\ref{fig:old-llms-performance}). It delivered perfect or near-perfect scores in the Computer Science (CS) Technologies category and demonstrated notably high accuracy in CS Fundamentals, particularly in the 2023 and 2024 exams. These results suggest that Gemini excels in algorithmic reasoning and technical problem-solving.

Model o1 also demonstrated strong and consistent performance, especially in Mathematics and CS Technologies. In both 2022 and 2024, it either matched or outperformed other models in these areas, highlighting its reliability across different exam editions. The o3-mini-high model performed competitively, often trailing closely behind Gemini in several categories, though its performance dipped slightly in the 2024 CS Technologies section.

In contrast, Claude 3.7 Sonnet showed greater variability. While it performed well in CS Fundamentals in 2022 and 2024, its results in Mathematics -- particularly in 2022 -- were considerably lower compared to the other models. This suggests that Claude may face challenges with symbolic and numerical reasoning tasks, especially when contrasted with the stronger-performing LLMs in this study.

Instances in which all models answer a question incorrectly are rare, occurring only once in 2022, once in 2023, and twice in 2024. In the 2022 exam, Question 56 -- related to types of non-functional requirements -- was answered incorrectly by all models, three of them selected a product requirement instead of an organizational one. Two of the paper's authors agreed with the LLMs' choice. In 2023, all models selected the same incorrect answer (B) for Question 29 (see Figure~\ref{fig:poscomp-29-2023}). In 2024, Questions 16 (Digital Circuits) and 52 (Databases) received identical incorrect responses from all four models.

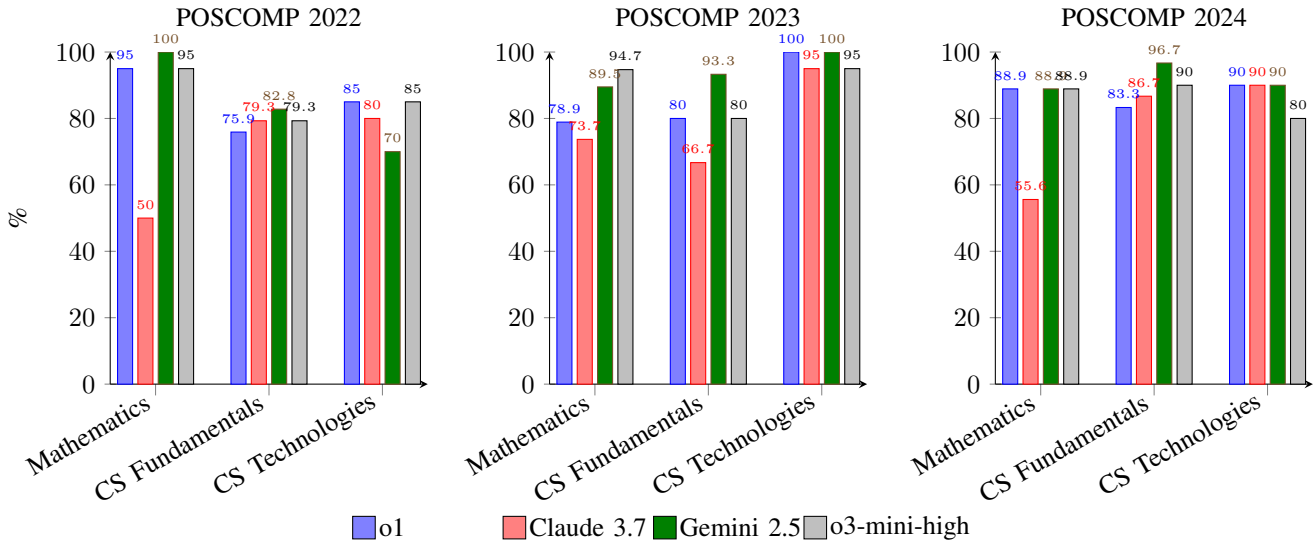
\begin{figure*}[ht]
\centering

\begin{minipage}{0.32\textwidth}
\centering
\begin{tikzpicture}
\begin{axis}[
    axis lines=left,
    ybar,
    bar width=.2cm,
    width=\linewidth,
    height=6cm,
    enlarge x limits=0.2,
    ylabel={\%},
    symbolic x coords={Mathematics, CS Fundamentals, CS Technologies},
    xtick=data,
    ymin=0, ymax=100,
    title={POSCOMP 2022},
    xticklabel style={rotate=30, anchor=east},
    nodes near coords,
    every node near coord/.append style={font=\tiny}
]
\addplot+[fill=blue!50] coordinates {(Mathematics, 95.0) (CS Fundamentals, 75.9) (CS Technologies, 85.0)};
\addplot+[fill=red!50] coordinates {(Mathematics, 50.0) (CS Fundamentals, 79.3) (CS Technologies, 80.0)};
\addplot+[fill=green!50!black] coordinates {(Mathematics, 100.0) (CS Fundamentals, 82.8) (CS Technologies, 70.0)};
\addplot+[fill=gray!50] coordinates {(Mathematics, 95.0) (CS Fundamentals, 79.3) (CS Technologies, 85.0)};
\end{axis}
\end{tikzpicture}
\end{minipage}
%
\begin{minipage}{0.32\textwidth}
\centering
\begin{tikzpicture}
\begin{axis}[
    axis lines=left,
    ybar,
    bar width=.2cm,
    width=\linewidth,
    height=6cm,
    enlarge x limits=0.2,
    ymin=0, ymax=100,
    symbolic x coords={Mathematics, CS Fundamentals, CS Technologies},
    xtick=data,
    title={POSCOMP 2023},
    xticklabel style={rotate=30, anchor=east},
    nodes near coords,
    every node near coord/.append style={font=\tiny}
]
\addplot+[fill=blue!50] coordinates {(Mathematics, 78.9) (CS Fundamentals, 80.0) (CS Technologies, 100.0)};
\addplot+[fill=red!50] coordinates {(Mathematics, 73.7) (CS Fundamentals, 66.7) (CS Technologies, 95.0)};
\addplot+[fill=green!50!black] coordinates {(Mathematics, 89.5) (CS Fundamentals, 93.3) (CS Technologies, 100.0)};
\addplot+[fill=gray!50] coordinates {(Mathematics, 94.7) (CS Fundamentals, 80.0) (CS Technologies, 95.0)};
\end{axis}
\end{tikzpicture}
\end{minipage}
%
\begin{minipage}{0.32\textwidth}
\centering
\begin{tikzpicture}
\begin{axis}[
    axis lines=left,
    ybar,
    bar width=.2cm,
    width=\linewidth,
    height=6cm,
    enlarge x limits=0.2,
    ymin=0, ymax=100,
    symbolic x coords={Mathematics, CS Fundamentals, CS Technologies},
    xtick=data,
    title={POSCOMP 2024},
    xticklabel style={rotate=30, anchor=east},
    nodes near coords,
    every node near coord/.append style={font=\tiny}
]
\addplot+[fill=blue!50] coordinates {(Mathematics, 88.9) (CS Fundamentals, 83.3) (CS Technologies, 90.0)};
\addplot+[fill=red!50] coordinates {(Mathematics, 55.6) (CS Fundamentals, 86.7) (CS Technologies, 90.0)};
\addplot+[fill=green!50!black] coordinates {(Mathematics, 88.9) (CS Fundamentals, 96.7) (CS Technologies, 90.0)};
\addplot+[fill=gray!50] coordinates {(Mathematics, 88.9) (CS Fundamentals, 90.0) (CS Technologies, 80.0)};
\end{axis}
\end{tikzpicture}
\end{minipage}

\vspace{-0.2cm}

\begin{center}
\parbox{0.9\textwidth}{
\centering
\begin{tikzpicture}[baseline, inner sep=0pt, outer sep=0pt]
\filldraw[fill=blue!50, draw=black] (0,0) rectangle (0.3,0.3); \node[anchor=west] at (0.35,0.15) {o1};
\filldraw[fill=red!50, draw=black] (2,0) rectangle (2.3,0.3); \node[anchor=west] at (2.35,0.15) {Claude 3.7};
\filldraw[fill=green!50!black, draw=black] (4,0) rectangle (4.3,0.3); \node[anchor=west] at (4.35,0.15) {Gemini 2.5};
\filldraw[fill=gray!50, draw=black] (6,0) rectangle (6.3,0.3); \node[anchor=west] at (6.35,0.15) {o3-mini-high};
\end{tikzpicture}
}
\end{center}

\vspace{-0.4cm}
\caption{Comparison of latest LLMs' performance on POSCOMP 2022, 2023, and 2024 by topic (in percentage).}
\label{fig:poscomp-results}
\end{figure*}

}

\subsubsection{Comparison with Students}

\revisao{
In the 2022 POSCOMP, average student scores were: 8.63/20 questions in Mathematics, 14/29 questions in Computer Science Fundamentals, and 6.58/20 questions in Computer Science Technologies. In contrast, all LLMs evaluated that year outperformed human averages across every topic. ChatGPT-4 stood out with 18/20 in Mathematics, 23/29 in CS Fundamentals, and 16/20 in CS Technologies. Gemini 1.0 also performed well with 14/20, 20/29, and 15/20, respectively. Mistral demonstrated consistent accuracy, especially in CS Technologies with 18/20 (90\%). Although Claude 3 had lower scores in Mathematics (11/20), it showed competitive results in the other two areas (18/29 and 15/20). Among the newer models, o1 and Gemini 2.5 both achieved perfect or near-perfect scores, such as 95\% to 100\% in Mathematics and over 82\% in CS Fundamentals. Claude 3.7 reached 82.8\% in CS Fundamentals but scored only 50\% in Mathematics, while o3-mini-high achieved a remarkable 90\% in CS Technologies, leading that category.

The 2023 POSCOMP results reinforced the gap between human and model performance. Students averaged 9.31/19 in Mathematics, 15.7/30 in CS Fundamentals, and 7.6/20 in CS Technologies. ChatGPT-4 led again with 17/19 in Mathematics, 24/30 in CS Fundamentals, and 18/20 in CS Technologies. Gemini 1.0 (13/19, 23/30, 17/20) and Mistral (14/19, 21/30, 15/20) followed closely. Claude 3 reached 9/19 in Mathematics and 22/30 in CS Fundamentals, placing among the top 10\%. Among the newer models, Gemini 2.5 and o3-mini-high reached 100\% accuracy in CS Technologies, while o1 and Claude 3.7 both surpassed 89\% in CS Fundamentals. Notably, Claude 3.7 improved over its earlier version, achieving 73.7\% in Mathematics and 66.7\% in CS Fundamentals.
ChatGPT-4, o1, Gemini 2.5, and o3-mini-high outperformed all human participants in the exam (the highest human score was 85.5\%). Among them, Gemini 2.5 achieved the best overall result, correctly answering 63 out of 69 questions (91.3\%).

In the 2024 POSCOMP, the average scores for students remained modest: 11.79/18 in Mathematics, 14.76/30 in CS Fundamentals, and 7.72/20 in CS Technologies. Once again, all LLMs surpassed human averages. o1 and o3-mini-high led in Mathematics with 16/18 (88.9\%), and Gemini 2.5 matched this score. In CS Fundamentals, Gemini 2.5 obtained the highest score with 29/30 (96.7\%), followed by o3-mini-high (27/30) and o1 (25/30). All models reached at least 80\% in CS Technologies, with o1, Claude 3.7, and Gemini 2.5 achieving 90\%. Even Claude 3.7, which had shown weaker performance in earlier editions, scored 16/18 in Mathematics and 26/30 in CS Fundamentals.
Notably, Gemini 2.5 Pro answered 63 out of 68 questions correctly (92.6\%) surpassing the highest score obtained by any student in the 2024 exam, which was 82.3\%. 
}

\subsection{Answers to Research Questions}

Next we answer the research questions.

\begin{itemize}%

\item[RQ$_{1}$] To what extent can ChatGPT-4 solve POSCOMP problems? \\
ChatGPT-4 achieved the highest scores among the tested models, correctly answering 57 out of 69 questions on the 2022 test and 59 out of 69 on the 2023 test. It excelled particularly in Mathematics and Computer Science Fundamentals. Its performance secured the best result in the 2023 exam, demonstrating its effectiveness in handling a broad range of topics, including complex technical questions.

\item[RQ$_{2}$] To what extent can Gemini 1.0 Advanced solve POSCOMP problems? \\
Gemini 1.0 Advanced also showed strong capability in solving POSCOMP problems, though slightly behind ChatGPT-4. It scored 49 out of 69 on the 2022 test and 53 out of 69 on the 2023 test. Gemini exhibited consistency across different subjects and maintained a robust performance in explanations, indicating a comprehensive approach to problem-solving.

\item[RQ$_{3}$] To what extent can Claude 3 Sonnet solve POSCOMP problems? \\
Claude 3 Sonnet had moderate success in solving POSCOMP problems, scoring 44 out of 69 in 2022 and 45 out of 69 in 2023. Despite demonstrating potential in explanations and understanding, Claude's performance was less consistent compared to ChatGPT-4 and Gemini. It encountered some difficulties in certain subjects like Mathematics, suggesting a potential area for improvement in tackling specific academic topics.

\item[RQ$_{4}$] To what extent can Le Chat Mistral solve POSCOMP problems? \\
\revisao{Le Chat Mistral demonstrated consistent but moderate performance across both the 2022 and 2023 POSCOMP exams, scoring 48 out of 69 (69.6\%) in 2022 and 50 out of 69 (72.5\%) in 2023. While its results were comparable to those of Gemini 1.0 and Claude 3, Mistral showed noticeable variability across topics, performing better in Computer Science Technologies than in Mathematics or Fundamentals.}

\revisao{
\item[\textbf{RQ$_{5}$}] To what extent do recent LLMs match or surpass human performance in POSCOMP exams? \\
The results indicate that recent LLMs not only match but  surpass human performance in POSCOMP exams. Across the 2022, 2023, and 2024 POSCOMP editions, all evaluated models outperformed average student scores in Mathematics, Computer Science Fundamentals, and Computer Science Technologies. More impressively, top-performing models such as ChatGPT-4, o1, Gemini 2.5, and o3-mini-high consistently outscored even the best human participants. For instance, Gemini 2.5 Pro Experimental scored 66 out of 69 questions in 2023 (95.6\%) and 63 out of 68 in 2024 (92.6\%), surpassing the top student scores of 84.05\% and 82.3\% in the respective years.
}

\end{itemize}

\section{Threats to Validity}
\label{sec:threats}

Some threats to validity can arise when using LLMs~\cite{threats-llms-icse-nier-2024}. The results are based on their performance on a specific exam, the POSCOMP, which means generalizing to other assessments or computer science contexts may not be straightforward. Variations in test formats, question types, or topics covered in other exams can lead to different performance outcomes. 
In our study, we aimed to minimize this threat by evaluating \revisao{three} editions of the POSCOMP.

The selection of four LLMs may not capture the full spectrum of current language model capabilities. Newer or differently configured models could deliver distinct results. 
We mitigated this threat by selecting four well-known and popular LLMs in the community.

The performance of LLMs heavily depends on the data used during training. Variations in the quality, quantity, and diversity of training data can significantly impact the results, potentially leading to an overestimation of the models' true ability to comprehend and process new information.
To help mitigate this threat, we applied metamorphic testing (Section~\ref{sec:metamorphic}) by introducing controlled changes to the wording of a few selected questions. ChatGPT-4 and Gemini were still able to answer all the modified questions correctly. 
\revisao{However, we acknowledge that applying metamorphic testing to a limited subset of questions does not fully eliminate the threat posed by variations in the training data.}

\section{Conclusions}
\label{sec:conclusao}

In this article, we evaluated the ability of four LLMs to solve questions from the 2022 and 2023 editions of the POSCOMP. Our findings reveal that large-scale language models (LLMs), particularly ChatGPT-4, perform well in addressing POSCOMP exam challenges. ChatGPT-4 consistently outperformed other models, achieving top scores in the 2023 edition. Its superior performance, particularly in Mathematics and Computer Science Fundamentals, underscores its potential as a formidable tool for students preparing for these competitive exams.
\revisao{
The evaluation of recent LLMs revealed a clear improvement in performance over earlier models. Gemini 2.5 Pro Experimental and o3-mini-high consistently achieved or exceeded 90\% accuracy across several topics, surpassing both average scores and the best-performing human participants. These findings highlight the rapid advancement of foundation models and their growing ability to tackle complex, domain-specific challenges in graduate-level assessments such as the POSCOMP.}

Other models, such as Gemini 1.0 Advanced and Le Chat Mistral, demonstrated problem-solving capabilities across various subjects but didn't match the achievements of ChatGPT-4. Claude 3 Sonnet showed promise in explanation and comprehension but lagged slightly behind, struggling particularly with Mathematics. Each LLM exhibited strengths and weaknesses in specific areas.

ChatGPT-4's success suggests that it has a more comprehensive training base and possibly more sophisticated mechanisms for interpreting and responding to diverse and complex queries compared to Gemini 1.0 Advanced. All models, however, encountered issues with incorrect interpretations, which were more pronounced and detrimental in Gemini 1.0. \revisao{These issues were notably less prevalent in the more recent LLMs.}

\revisao{
Incorporating multiple LLMs into the question development process can be a valuable strategy for problem setters. By observing how different models interpret and respond to each question, it becomes easier to identify issues such as ambiguity, lack of clarity, or incomplete information. This process not only supports the refinement of problem statements and answer options but also helps ensure that questions are well-structured and accessible to a broader range of test-takers. Additionally, discrepancies in model responses may indicate parts of a question that could be misinterpreted, prompting targeted revisions before finalizing the exam.
}

\revisao{
For future work, we plan to broaden our evaluation by incorporating additional LLMs, such as DeepSeek, to further investigate model diversity and performance. We also aim to explore the impact of advanced prompt engineering techniques~\cite{prompts,prompt-techniques}, including few-shot and chain-of-thought prompting, on the models’ ability to solve complex academic tasks. This extended analysis will deepen our understanding of how LLMs can be effectively leveraged in educational contexts. Furthermore, we intend to expand our use of metamorphic testing by applying a larger and more diverse set of systematically transformed questions. While our initial experiments introduced controlled alterations to assess robustness, future work will explore more transformation strategies that preserve semantic meaning while varying surface-level features. This will allow us to better examine model sensitivity to input variation and further investigate issues related to memorization, generalization, and training data overlap.

As LLMs continue to evolve, we anticipate that many of the current limitations will be reduced or eliminated. For instance, future models may offer support for significantly larger context windows and more flexible input formats, enabling the evaluation of full-length PDF exams without the need to divide them into smaller segments. Such advancements would not only improve model performance by preserving context across multiple questions, but also enhance the applicability of LLMs in educational assessment scenarios.
}

\section*{Acknowledgements}
We thank the anonymous reviewers for their valuable feedback and suggestions, which helped to improve the quality of this article.
This work is partially supported by CNPq and CAPES grants.



\begin{thebibliography}{10}
\providecommand{\url}[1]{#1}
\csname url@samestyle\endcsname
\providecommand{\newblock}{\relax}
\providecommand{\bibinfo}[2]{#2}
\providecommand{\BIBentrySTDinterwordspacing}{\spaceskip=0pt\relax}
\providecommand{\BIBentryALTinterwordstretchfactor}{4}
\providecommand{\BIBentryALTinterwordspacing}{\spaceskip=\fontdimen2\font plus
\BIBentryALTinterwordstretchfactor\fontdimen3\font minus \fontdimen4\font\relax}
\providecommand{\BIBforeignlanguage}[2]{{%
\expandafter\ifx\csname l@#1\endcsname\relax
\typeout{** WARNING: IEEEtran.bst: No hyphenation pattern has been}%
\typeout{** loaded for the language `#1'. Using the pattern for}%
\typeout{** the default language instead.}%
\else
\language=\csname l@#1\endcsname
\fi
#2}}
\providecommand{\BIBdecl}{\relax}
\BIBdecl

\bibitem{poscomp}
{Sociedade Brasileira de Computação}, ``{POSCOMP},'' \url{https://www.sbc.org.br}, 2025.

\bibitem{Goodfellow-et-al-2016}
I.~Goodfellow, Y.~Bengio, and A.~Courville, \emph{Deep Learning}.\hskip 1em plus 0.5em minus 0.4em\relax MIT Press, 2016.

\bibitem{attention-is-all-you-need}
A.~Vaswani, N.~Shazeer, N.~Parmar, J.~Uszkoreit, L.~Jones, A.~N. Gomez, L.~Kaiser, and I.~Polosukhin, ``Attention is all you need,'' in \emph{Advances in Neural Information Processing Systems}, 2017, pp. 5998--6008.

\bibitem{DBLP:journals/toce/Mendonca24}
N.~C. Mendon{\c{c}}a, ``Evaluating {ChatGPT-4} vision on brazil's national undergraduate computer science exam,'' \emph{{ACM} Transactions on Computing Education}, vol.~24, no.~3, pp. 1--56, 2024.

\bibitem{pires}
\BIBentryALTinterwordspacing
R.~Pires, T.~S. Almeida, H.~Q. Abonizio, and R.~F. Nogueira, ``Evaluating gpt-4's vision capabilities on brazilian university admission exams,'' vol. abs/2311.14169, 2023. [Online]. Available: \url{https://doi.org/10.48550/arXiv.2311.14169}
\BIBentrySTDinterwordspacing

\bibitem{chain-of-thought}
J.~Wei, X.~Wang, D.~Schuurmans, M.~Bosma, B.~Ichter, F.~Xia, E.~H. Chi, Q.~V. Le, and D.~Zhou, ``Chain-of-thought prompting elicits reasoning in {L}arge {L}anguage {M}odels,'' in \emph{Advances in Neural Information Processing Systems}, 2022.

\bibitem{nunes}
\BIBentryALTinterwordspacing
D.~Nunes, R.~Primi, R.~Pires, R.~de~Alencar~Lotufo, and R.~F. Nogueira, ``Evaluating {GPT-3.5} and {GPT-4} models on brazilian university admission exams,'' vol. abs/2303.17003, 2023. [Online]. Available: \url{https://doi.org/10.48550/arXiv.2303.17003}
\BIBentrySTDinterwordspacing

\bibitem{gpt}
{OpenAI}, ``{ChatGPT},'' \url{https://chat.openai.com}, 2024.

\bibitem{gemini}
{Google}, ``{Gemini},'' \url{https://gemini.google.com}, 2024.

\bibitem{claude}
{Anthropic}, ``{Claude},'' \url{https://claude.ai}, 2024.

\bibitem{mistral}
{Mistral}, ``{Le Chat Mistral},'' \url{https://chat.mistral.ai/chat}, 2024.


\bibitem{zhang}
\BIBentryALTinterwordspacing
X.~Zhang, C.~Li, Y.~Zong, Z.~Ying, L.~He, and X.~Qiu, ``Evaluating the performance of large language models on {GAOKAO} benchmark,'' vol. abs/2305.12474, 2023. [Online]. Available: \url{https://doi.org/10.48550/arXiv.2305.12474}
\BIBentrySTDinterwordspacing

\bibitem{grima}
F.~Guillen-Grima, S.~Guillen-Aguinaga, L.~Guillen-Aguinaga, R.~Alas-Brun, L.~Onambele, W.~Ortega, R.~Montejo, E.~Aguinaga-Ontoso, P.~Barach, and I.~Aguinaga-Ontoso, ``Evaluating the efficacy of chatgpt in navigating the spanish medical residency entrance examination (mir): Promising horizons for ai in clinical medicine,'' \emph{Clinics and Practice}, vol.~13, no.~6, pp. 1460--1487, 2023.

\bibitem{bommaritoM}
\BIBentryALTinterwordspacing
M.~J.~B. II and D.~M. Katz, ``{GPT} takes the bar exam,'' vol. abs/2212.14402, 2022. [Online]. Available: \url{https://doi.org/10.48550/arXiv.2212.14402}
\BIBentrySTDinterwordspacing

\bibitem{BommaritoJ}
\BIBentryALTinterwordspacing
J.~Bommarito, M.~J.~B. II, D.~M. Katz, and J.~Katz, ``{GPT} as knowledge worker: {A} zero-shot evaluation of {(AI)CPA} capabilities,'' vol. abs/2301.04408, 2023. [Online]. Available: \url{https://doi.org/10.48550/arXiv.2301.04408}
\BIBentrySTDinterwordspacing

\bibitem{joshi}
I.~Joshi, R.~Budhiraja, H.~Dev, J.~Kadia, M.~O. Ataullah, S.~Mitra, H.~D. Akolekar, and D.~Kumar, ``Chatgpt in the classroom: An analysis of its strengths and weaknesses for solving undergraduate computer science questions,'' in \emph{Proceedings of the 55th {ACM} Technical Symposium on Computer Science Education}.\hskip 1em plus 0.5em minus 0.4em\relax {ACM}, 2024, pp. 625--631.

\bibitem{espejel}
J.~{López Espejel}, E.~H. Ettifouri, M.~S. {Yahaya Alassan}, E.~M. Chouham, and W.~Dahhane, ``{GPT-3.5, GPT-4, or BARD}? evaluating {LLM}s reasoning ability in zero-shot setting and performance boosting through prompts,'' \emph{Natural Language Processing Journal}, vol.~5, p. 100032, 2023.

\bibitem{toyama}
Y.~Toyama, A.~Harigai, M.~Abe, M.~Nagano, M.~Kawabata, Y.~Seki, and K.~Takase, ``Performance evaluation of {ChatGPT, GPT-4, and Bard} on the official board examination of the japan radiology society,'' \emph{Japanese Journal of Radiology}, vol.~42, pp. 201--207, 2024.

\bibitem{intrator}
\BIBentryALTinterwordspacing
Y.~Intrator, M.~Halfon, R.~Goldenberg, R.~Tsarfaty, M.~Eyal, E.~Rivlin, Y.~Matias, and N.~Aizenberg, ``Breaking the language barrier: Can direct inference outperform pre-translation in multilingual {LLM} applications?'' vol. abs/2403.04792, 2024. [Online]. Available: \url{https://doi.org/10.48550/arXiv.2403.04792}
\BIBentrySTDinterwordspacing

\bibitem{prompt-techniques}
{DAIR.AI}, ``{Prompt Engineering Guide},'' \url{https://www.promptingguide.ai/techniques}, 2025.

\bibitem{prompts}
P.~Liu, W.~Yuan, J.~Fu, Z.~Jiang, H.~Hayashi, and G.~Neubig, ``Pre-train, prompt, and predict: A systematic survey of prompting methods in natural language processing,'' \emph{ACM Computing Surveys (CSUR)}, vol.~55, no.~9, pp. 1--35, 2023.

\bibitem{threats-llms-icse-nier-2024}
J.~Sallou, T.~Durieux, and A.~Panichella, ``Breaking the silence: the threats of using llms in software engineering,'' in \emph{ACM/IEEE 46th International Conference on Software Engineering - New Ideas and Emerging Results}.\hskip 1em plus 0.5em minus 0.4em\relax ACM/IEEE, 2024.

\bibitem{metamorphic-testing}
L.~Applis, A.~Panichella, and R.~Marang, ``Searching for quality: Genetic algorithms and metamorphic testing for software engineering {ML},'' in \emph{Proceedings of the Genetic and Evolutionary Computation Conference}.\hskip 1em plus 0.5em minus 0.4em\relax {ACM}, 2023, pp. 1490--1498.

\bibitem{metamorphic-testing-2}
T.~Y. Chen, F.~Kuo, H.~Liu, P.~Poon, D.~Towey, T.~H. Tse, and Z.~Q. Zhou, ``Metamorphic testing: {A} review of challenges and opportunities,'' \emph{{ACM} Computing Surveys}, vol.~51, no.~1, pp. 4:1--4:27, 2018.

\end{thebibliography}
\end{document}